\documentclass[conference]{IEEEtran}
\IEEEoverridecommandlockouts

\usepackage[left=0.75in,
            right=0.75in, 
            top=0.75in,
            bottom=0.75in]{geometry}
            
\usepackage{cite}
\usepackage{amsmath,amssymb,amsfonts}
\usepackage{algorithmic}
\usepackage{graphicx}
\usepackage{rotating}
\usepackage{multirow}
\usepackage{sansmath}
\usepackage{textcomp}
\usepackage{xcolor} 
\def\BibTeX{{\rm B\kern-.05em{\sc i\kern-.025em b}\kern-.08em
    T\kern-.1667em\lower.7ex\hbox{E}\kern-.125emX}}
\usepackage{times}
\usepackage{soul}
\usepackage{url}
\usepackage[hidelinks]{hyperref}
\usepackage[utf8]{inputenc}
\usepackage[small]{caption}
\usepackage{graphicx}
\usepackage{amsmath}

\usepackage{amsthm}
\usepackage{booktabs}
\usepackage{algorithm}
\usepackage{algorithmic}
\usepackage{multirow}
\usepackage{array}
\usepackage{makecell}
\usepackage[switch]{lineno}
\usepackage{tikz}
\usepackage[edges]{forest}
\definecolor{hidden-draw}{RGB}{205, 44, 36}
\definecolor{hidden-blue}{RGB}{194,232,247}
\definecolor{hidden-orange}{RGB}{243,202,120}
\definecolor{hidden-yellow}{RGB}{242,244,193}
\definecolor{tree-level-1}{RGB}{245,20,85}
\definecolor{tree-level-2}{RGB}{246,86,118}
\definecolor{tree-level-3}{RGB}{248,177,193}
\definecolor{tree-leaf}{RGB}{176,230,198}

\definecolor{Self}{RGB}{255,0,128}
\definecolor{Ensemble}{RGB}{0,127,255}
\definecolor{Iterative}{RGB}{153,51,255}

\definecolor{exemplar1}{RGB}{136,98,148}
\definecolor{exemplar2}{RGB}{148,210,242}
\definecolor{knowledge1}{RGB}{249,219,152}
\definecolor{knowledge2}{RGB}{255,245,220}

\definecolor{skyblue}{RGB}{86, 180, 233}
\definecolor{bluishgreen}{RGB}{0, 158, 115}
\definecolor{yellow}{RGB}{240, 228, 66}
\definecolor{blue}{RGB}{0, 114, 178}
\definecolor{vermillion}{RGB}{213, 94, 0}
\definecolor{reddishpurple}{RGB}{204, 121, 167}
\definecolor{saffron}{RGB}{244, 196, 48}

\usepackage{wrapfig}
\usepackage{enumitem}
\begin{document}

\title{Knowledge-informed Molecular Learning: \newline
A Survey on Paradigm Transfer
}

\author{
   \IEEEauthorblockN{Yin Fang\textsuperscript{$\clubsuit$}, Zhuo Chen\textsuperscript{$\clubsuit$}, Xiaohui Fan\textsuperscript{$\heartsuit$}, Ningyu Zhang\textsuperscript{$\clubsuit$,$\spadesuit$,*\thanks{* Corresponding Author}}}
   \vspace{0.3cm}
   \IEEEauthorblockA{\textit{\textsuperscript{$\clubsuit$}College of Computer Science and Technology, Zhejiang University}}
   \IEEEauthorblockA{\textit{\textsuperscript{$\spadesuit$}School of Software Technology, Zhejiang University}}
   \IEEEauthorblockA{\textit{\textsuperscript{$\heartsuit$}College of Pharmaceutical Sciences, Zhejiang University}} \\
   \small \{\texttt{fangyin}, \texttt{zhuo.chen}, \texttt{fanxh}, \texttt{zhangningyu} \}\texttt{@zju.edu.cn}\\
}

\maketitle
\thispagestyle{plain}
\pagestyle{plain}

\begin{abstract}
Machine learning, notably deep learning, has significantly propelled molecular investigations within the biochemical sphere. 
Traditionally, modeling for such research has centered around a handful of paradigms. For instance, the prediction paradigm is frequently deployed for tasks such as molecular property prediction. 
To enhance the generation and decipherability of purely data-driven models, scholars have integrated biochemical domain knowledge into these molecular study models. 
This integration has sparked a surge in paradigm transfer, which is solving one molecular learning task by reformulating it as another one.
With the emergence of Large Language Models, these paradigms have demonstrated an escalating trend towards harmonized unification.
In this work, we delineate a literature survey focused on knowledge-informed molecular learning from the perspective of paradigm transfer.
We classify the paradigms, scrutinize their methodologies, and dissect the contribution of domain knowledge. 
Moreover, we encapsulate prevailing trends and identify intriguing avenues for future exploration in molecular learning.  
\end{abstract}


\section{Introduction}\label{intro}

Just as the syntax of natural languages enforces a grammatical structure that facilitates the connection between words in both sequential and hierarchical ways through sentence structures and parsing graphs, biological symbols also amalgamate in precise structural manners represented by either molecular strings or molecular graphs. 
As a result, machine learning has witnessed a proliferation in its applications spanning diverse scientific disciplines, such as biology and chemistry. 

The primary areas of focus in the molecule field include molecular property prediction, molecule-molecule interaction prediction, molecule generation and optimization, reaction and retrosynthesis. 
Specifically, the aim of molecular property prediction is to accurately forecast the physical, chemical, or biological properties of molecules, thereby expediting efficient drug design and discovery. 
Molecule-molecule interaction prediction aids in evaluating the safety and potency of multiple drug amalgamations, mitigating the peril of adverse reactions. 
Molecule generation and optimization strive to fabricate novel chemical compounds or augment existing ones with desired properties, enabling a more thorough exploration of the expansive chemical space.
Chemical reaction and retrosynthesis facilitate the design of innovative reactions and syntheses, enhancing our comprehension of the reactivity and synthetic feasibility of molecules for applications in chemistry and pharmaceuticals. 
Machine learning models grapple with vast quantities of data, deciphering intricate relationships between molecular properties, interactions, and reactions, leading to more precise predictions and a more profound understanding of the molecular world.

However, purely data-driven models, while effective, bear several shortcomings. 
\textbf{Firstly}, these models may lack a holistic understanding of physical and chemical processes, potentially hampering their accuracy in areas of the chemical space that are underrepresented in the training data. 
\textbf{Secondly}, overfitting to the training data can induce poor generalization to novel data.
\textbf{Thridly}, these models can be complex and difficult to interpret, making the elucidation of the logic behind predictions a formidable challenge.
The infusion of accumulated expert insights (i.e., domain knowledge) can bolster the resilience of molecular learning models, curbing the influence of noisy data. Hence, the integration of domain knowledge is pivotal in molecular learning.

\begin{figure}[t] 
\centering 
 \includegraphics[width=1\columnwidth]{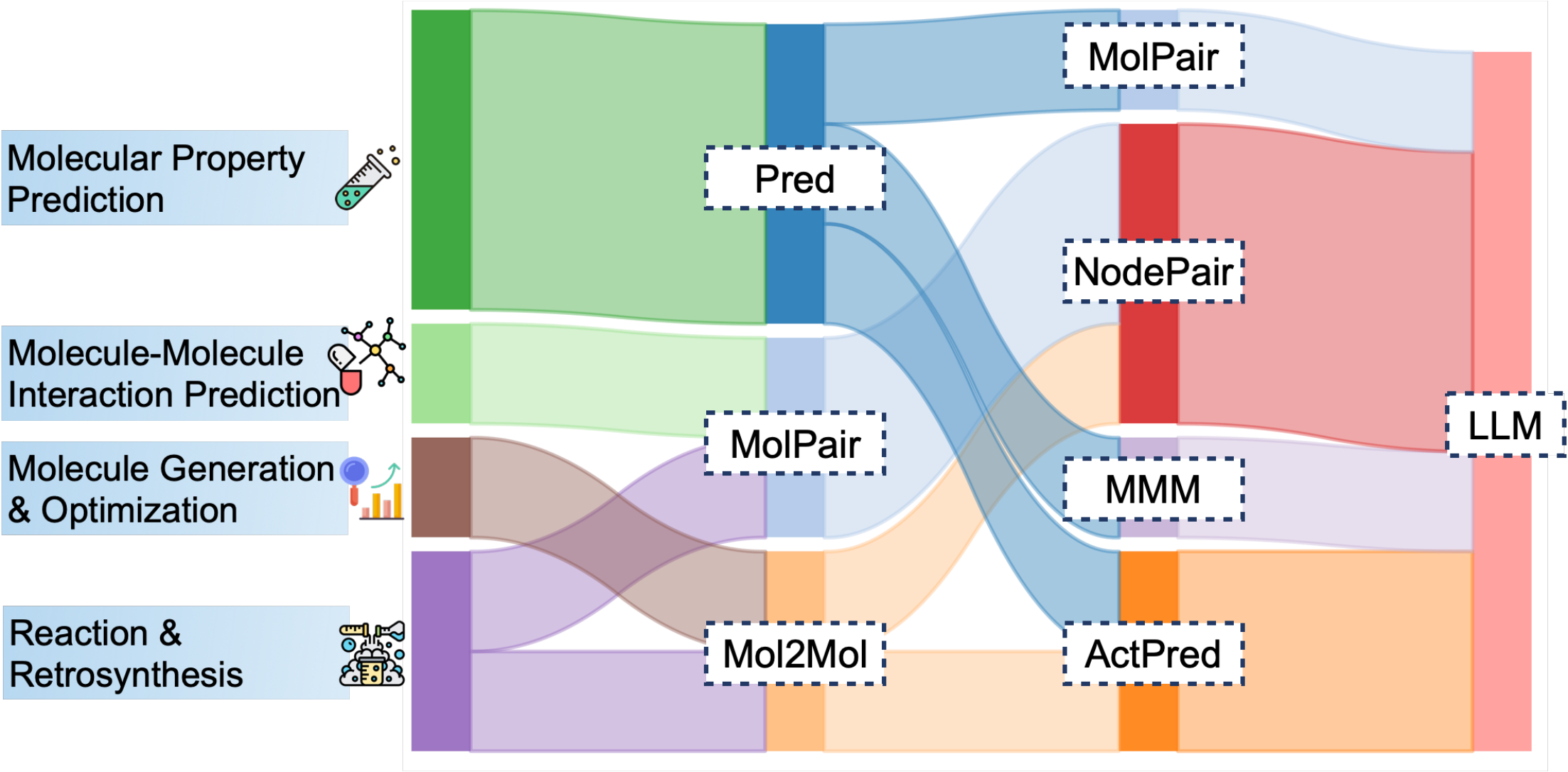}
    \caption{A Sankey diagram illustrating paradigm transfer (right) of different molecular tasks (left).}
\label{fig:sankey}
\vspace{-0.5cm}
\end{figure}

Steered by domain knowledge, the molecular learning paradigm has undergone a transformation, as depicted in Figure~\ref{fig:sankey}. 
A paradigm denotes a comprehensive machine learning framework designed for a specific set of tasks.
Various paradigms typically encompass distinct model architectures and inputs/outputs, heavily contingent on the nature of the tasks in question. 
Molecular learning tasks can be classified into several paradigms, including Prediction (\textsc{Pred}), Molecule Pairing (\textsc{MolPair}), Node Pairing (\textsc{NodePair}), Masked Molecule Model (\textsc{Mmm}), Molecule-to-Molecule (\textsc{Mol2Mol}), Action Prediction (\textsc{ActPred}), and at the forefront, the Large Language Model (\textsc{Llm}), as elucidated in this paper.
To transition between paradigms, the input data must initially be reformatted into the suitable format, followed by processing by the new paradigm to solve the desired task.
In recent years, there has been an escalating interest in generalizing paradigms to diverse tasks, guided by domain knowledge, resulting in increased success and recognition within the molecular learning community.



These knowledge-informed paradigm transfers interspersed in various molecular studies have not been systematically reviewed and analyzed.
In this paper, we strive to encapsulate recent developments and trends in this line of research. 
This paper is structured as follows. 
In \S ~\ref{paradigm}, we furnish formal definitions of the seven paradigms and introduce their representative tasks and instance models. 
In \S ~\ref{paradigm transfer}, we scrutinize how domain knowledge directs transfers across various molecular learning tasks.
In \S ~\ref{conclusion}, we conclude with a discourse on current trends and prospective directions in this field.

\section{Molecular Learning Paradigms}\label{paradigm}

\begin{figure*}[!ht] 
\centering 
\includegraphics[width=0.95\textwidth]{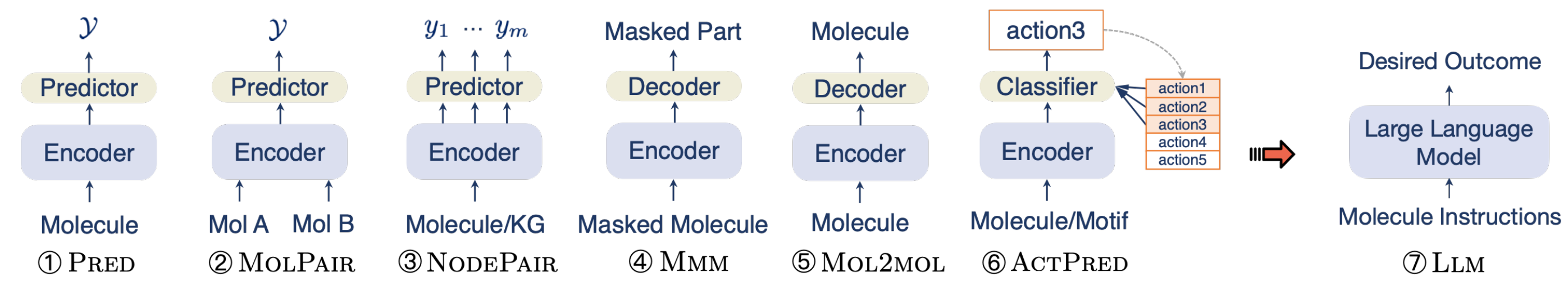}
\caption{Illustration of the seven mainstream paradigms in molecular learning.} 
\vspace{-0.5cm}
\label{fig:paradigm}
\end{figure*}

In this section, we delineate the prevailing paradigms employed in molecular learning tasks (Figure~\ref{fig:paradigm}), encompassing their representative tasks and associated models. Specifically, a paradigm serves as a universal framework for accommodating datasets or tasks of a particular format~\cite{DBLP:journals/ijautcomp/SunLQH22}. 

\subsection{Prediction (\textsc{Pred})}

The \textsc{Pred} paradigm aims to predict a pre-determined label or a probability for a specified molecule.
It is extensively utilized in foundational molecular tasks such as predicting molecular properties~\cite{DBLP:conf/ijcai/SongZNFLY20,DBLP:journals/corr/abs-2012-11175} and estimating probabilities of candidate products/reaction rules~\cite{coley2017prediction,DBLP:conf/nips/JinCBJ17,coley2017computer}. 
\textsc{Pred} is accomplished by directing the input into a deep neural network-based encoder to extract a task-specific embedding, which is subsequently fed into a predictor module to generate the output value, i.e., 
\begin{equation}
	\mathcal{Y} = \textsc{Pre}(\textsc{Enc}(\mathcal{X})),
\end{equation}

where $\mathcal{Y}$ signifies either a value (for regression tasks) or a one-hot/multi-hot vector (for classification tasks). 
The input $\mathcal{X}$ could be a molecular graph or its SMILES~\cite{DBLP:journals/jcisd/WeiningerWW89} or SELFIES~\cite{DBLP:journals/mlst/KrennHNFA20} sequence.
$\textsc{Enc}(\cdot)$ and $\textsc{Pre}(\cdot)$ symbolize the encoder and predictor, respectively.
$\textsc{Enc}(\cdot)$ could use a GNN~\cite{xiong2019pushing}, RNN~\cite{DBLP:conf/bcb/0005WZH17}, or Transformer~\cite{DBLP:journals/corr/abs-2107-08773} to manage disparate inputs.
Following encoding, a pooling layer summarizes the molecule representation, and $\textsc{Pre}(\cdot)$, typically an MLP or a linear layer, generates predictions.

\subsection{Molecule Pairing (\textsc{MolPair})}


The \textsc{MolPair} paradigm holds paramount importance in various molecular learning tasks, such as predicting molecule-molecule interaction prediction~\cite{DBLP:conf/aaai/HuangXHGS20,nyamabo2021drug} and conducting molecular contrastive learning in self-supervised molecular representation learning~\cite{DBLP:journals/corr/abs-2106-04509,DBLP:journals/corr/abs-2102-10056,DBLP:conf/aaai/FangZYZD0Q0FC22}. 
The objective is to predict the relevance between two molecules while capturing both the features of the molecules and their fine-grained interaction. 
This paradigm can be expressed as
\begin{equation}
	\mathcal{Y} = \textsc{Pre}(\textsc{Enc}(\mathcal{X}_a, \mathcal{X}_b)),
\end{equation}
where $\mathcal{X}_a$ and $\mathcal{X}_b$ are two molecules under prediction, and $\mathcal{Y}$ can be discrete (e.g., indicating whether one molecule exerts a pharmacological effect on the other) or continuous (e.g., representing the similarity between two drugs). 
The two molecules can be encoded individually~\cite{DBLP:journals/ws/AbdelazizFHZS17} or in conjunction~\cite{DBLP:conf/www/WangMCW21}, and then fed into $\textsc{Pre}(\cdot)$ to capture their interaction.

\subsection{Node Pairing (\textsc{NodePair})}


The \textsc{NodePair} paradigm is widely used in graph structure prediction. 
Contrasting the Molecule Pairing (\textsc{MolPair}) paradigm, \textsc{NodePair} concentrates on investigating interactions between atoms within a molecular graph or entities in a biochemical knowledge graph (KG).  
Traditional neural network-based node pairing models employ an encoder to capture the representation of each node within the graph, and a predictor to evaluate the relevance of the node pairs,
\begin{equation}
    y_1, \dots, y_m = \textsc{Pre}(\textsc{Enc}(\mathcal{X})).
\end{equation}
where $y_1, \dots, y_m$ denote the labels of the $m$ node pairs to be predicted in the molecule $\mathcal{X}$.
The encoder is typically embodied as a graph neural network (GNN), and the predictor as a multi-layer perceptron (MLP).
For molecular graphs, the \textsc{NodePair} problem can be conceived as a binary classification task, aiming to predict the presence of a chemical bond between two atoms~\cite{DBLP:conf/nips/YanDZZY0H20}. 
In the context of biochemical KGs, it metamorphoses into a multi-class classification problem, with the objective of predicting the probability of an edge between two entities~\cite{DBLP:conf/ijcai/LinQWMZ20,DBLP:conf/ijcai/LyuGTLZZ21}.

\subsection{Masked Molecule Model (\textsc{Mmm})}


The \textsc{Mmm} paradigm in molecular learning predicts a substructure's likelihood in a molecule~\cite{DBLP:conf/iclr/HuLGZLPL20} or its attributes~\cite{DBLP:conf/nips/RongBXX0HH20,zhang2021motif}. 
Although these two tasks are somewhat distinct, they share the aim of estimating the probability of a masked section given context. 
The self-supervised fashion of \textsc{Mmm} enables its adoption as a pretext task to pre-train models on large-scale unlabeled molecular data. Typically, \textsc{Mmm} can be formulated as
\begin{equation}
    \bar{\mathcal{X}} = \textsc{Dec}(\textsc{Enc}(\tilde{\mathcal{X}})),
\end{equation}
where $\tilde{\mathcal{X}}$ is a corrupted version of the original molecule $\mathcal{X}$, with certain substructures or attributes masked, and $\bar{\mathcal{X}}$ represents the masked part predicted.
For molecular graphs, $\textsc{Enc}(\cdot)$ can be a GNN-based or Transformer-based model~\cite{DBLP:conf/nips/RongBXX0HH20}, and $\textsc{Dec}(\cdot)$ typically an MLP. 
For molecular strings, $\textsc{Enc}(\cdot)$ can be a bidirectional Transformer, and $\textsc{Dec}(\cdot)$ an autoregressive Transformer.

\subsection{Molecule-to-Molecule (\textsc{Mol2Mol})}


The \textsc{Mol2Mol} paradigm focuses on tasks like molecular generation, drug design, and molecule optimization, aiming to generate a new molecule from a given one. 
There are mainly two versions of \textsc{Mol2Mol}. The first is commonly implemented using an autoencoder-decoder architecture~\cite{DBLP:conf/iclr/DaiTDSS18}, where an encoder network transforms the molecules into fixed-dimensional vectors, and a decoder network converts these vectors back into molecules:
 \begin{equation}
		\mathcal{X}' = \textsc{Dec}(\textsc{Enc}(\mathcal{X})).
\end{equation}
Here, $\mathcal{X}$ refers to the original molecule or a substructure, and $\mathcal{X}'$ is the target molecule. The second version, working on molecular strings, is typically expressed as:
\begin{equation}
    y_1, \dots, y_m = \textsc{Dec}(\textsc{Enc}(x_1, \dots, x_n)),
\end{equation}
with $\textsc{Enc}(\cdot)$ and $\textsc{Dec}(\cdot)$ often instantiated as RNNs~\cite{popova2018deep} or Transformers~\cite{DBLP:journals/natmi/WangHWWWJLZYHCC21}. 
Notably, the input and output sequence lengths can differ and the decoder, using either the previous output during inference or ground truth during training, may be more complex.

\subsection{Action Prediction (\textsc{ActPred})}


The \textsc{ActPred} paradigm optimizes molecular properties by selecting actions that transform an initial molecular state to a terminal one.
Widely employed in molecular generation and optimization, the candidate action pool generally includes molecular modifications such as adding atoms, edges, or motifs~\cite{yang2021hit}, and reactant selection, where the product molecule becomes the input in the next step~\cite{DBLP:conf/icml/GottipatiSNPWLL20}. 
This paradigm is denoted as:
\begin{equation}
    \mathcal{A} = \textsc{Cls}(\textsc{Enc}(\mathcal{X}), \mathcal{C}),
\end{equation}
where $\mathcal{A} = \{a_1, \dots, a_m\}$ represents a sequence of actions, and $\mathcal{C} = \{c_0, \dots, c_{m-1}\}$ is a sequence of states. 
At each time step $t$, an action $a_t$ is predicted based on the input and the current state $c_{t-1}$, which might contain previous actions and reaction template~\cite{DBLP:conf/icml/GottipatiSNPWLL20}.

\subsection{Large Language Model (\textsc{Llm})}
The advent of Large Language Models (LLMs) such as GPT-4~\cite{DBLP:journals/corr/abs-2303-08774}, LLaMA~\cite{DBLP:journals/corr/abs-2302-13971}, FLAN~\cite{DBLP:conf/iclr/WeiBZGYLDDL22}, and GLM~\cite{DBLP:journals/corr/abs-2210-02414} has ushered in a transformative era in Natural Language Processing (NLP). 
These models, each endowed with billions of parameters, undergo meticulous training on vast text corpora, exhibiting an exceptional ability to generate text akin to human language, while also unraveling subtle distinctions in various expressions and nuances.
In light of this, the \textsc{Llm} paradigm capitalizes on the directive capacities of LLMs to decode, produce, and enhance molecular structures and properties.

Within the \textsc{Llm} paradigm, molecular task instructions, often encoded as text sequences, serve as the input. For example, ``\textbf{Create a molecule that satisfies the conditions outlined in the description:} \textit{The molecule appears as a yellow or red crystalline solid or powder ...}''
Such instructions encompass a spectrum of requirements for molecular design, optimization, and prediction. 
Subsequently, the formidable language comprehension prowess of the LLMs is harnessed to decipher instructions and provide an output that aligns with the requisites of the instruction.

The \textsc{Llm} paradigm encapsulates the fusion of human language and molecular science, transmuting textual guidance into tangible molecular realities. 
Blending an extensive reservoir of knowledge, it unifies a diverse array of paradigms into a cohesive framework, ushering in a novel era of computational exploration in the realm of molecular studies.

\section{Knowledge-informed Paradigm Transfer}\label{paradigm transfer}
Recent studies have revealed that models under certain paradigms can generalize well on tasks with other paradigms, guided by various forms of domain knowledge. 
In this section, we survey the knowledge-informed paradigm transfers that are prevalent in typical molecular learning tasks, including molecular property prediction, molecule-molecule interaction prediction, molecule generation and optimization, reaction and retrosynthesis prediction. 
We also examine the role of domain knowledge in these transfers. 
Furthermore, we explore how the \textsc{Llm} paradigm integrates the above molecular learning tasks.
A taxonomy of these transfers is illustrated in Figure~\ref{fig:paradigm transfer}.

\tikzstyle{my-box}=[
    rectangle,
    draw=hidden-draw,
    rounded corners,
    text opacity=1,
    minimum height=1.5em,
    minimum width=5em,
    inner sep=2pt,
    align=center,
    fill opacity=.5,
]
\tikzstyle{leaf}=[my-box, minimum height=1.5em,
    fill=hidden-orange!60, text=black, align=left,font=\scriptsize,
    inner xsep=2pt,
    inner ysep=4pt,
]
\tikzstyle{para}=[my-box, minimum height=1.5em,
    fill=hidden-draw!60, text=black, align=left,font=\scriptsize,
    inner xsep=2pt,
    inner ysep=4pt,
]
\begin{figure*}[!t]
    \centering
    \resizebox{\textwidth}{!}{
        \begin{forest}
            forked edges,
            for tree={
                grow=east,
                reversed=true,
                anchor=base west,
                parent anchor=east,
                child anchor=west,
                base=center,
                font=\small,
                rectangle,
                draw=hidden-draw,
                rounded corners,
                align=center,
                minimum width=4em,
                edge+={darkgray, line width=1pt},
                s sep=3pt,
                inner xsep=2pt,
                inner ysep=3pt,
                ver/.style={rotate=90, child anchor=north, parent anchor=south, anchor=center},
            },
            where level=1{text width=6.7em,font=\scriptsize,}{},
            where level=2{text width=3em,font=\scriptsize,}{},
            where level=3{text width=10em,font=\scriptsize,}{},
            where level=4{text width=3em,font=\scriptsize,}{},
            where level=5{text width=22em,font=\scriptsize,}{},
            [
                Molecular Learning, ver
                [
                    Molecular Property\\ Prediction
                    [
                        \textsc{Pred}
                        , para, text width=4em
                        [
                            NeuralFP~\cite{DBLP:conf/nips/DuvenaudMABHAA15}{,} seq2seqFP~\cite{DBLP:conf/bcb/0005WZH17}{,}\\
                            MPNN~\cite{DBLP:conf/icml/GilmerSRVD17}{,}
                            DMPNN~\cite{DBLP:journals/jcisd/YangSJCEGGHKMPS19}{,}\\
                            CMPNN~\cite{DBLP:conf/ijcai/SongZNFLY20}{,}
                            CoMPT~\cite{DBLP:journals/corr/abs-2107-08773}\\
                            , leaf
                            [
                                \textsc{MolPair}
                                , para, text width=4em
                                [
                                    MoCL~\cite{DBLP:journals/corr/abs-2106-04509}{,}
                                    CKGNN~\cite{DBLP:journals/corr/abs-2103-13047}{,}
                                    KCL~\cite{DBLP:conf/aaai/FangZYZD0Q0FC22}{,}
                                    KANO~\cite{fang2023knowledge}
                                    , leaf, text width=16.5em
                                ]
                            ]
                            [
                                \textsc{Mmm}
                                , para, text width=4em
                                [
                                    Hu et al.~\cite{DBLP:conf/iclr/HuLGZLPL20}{,}
                                    GROVER~\cite{DBLP:conf/nips/RongBXX0HH20}{,}
                                    Mole-BERT~\cite{DBLP:conf/iclr/XiaZHG0LLL23}
                                    , leaf, text width=15em
                                ]
                            ]
                            [
                                \textsc{ActPred}
                                , para, text width=3.5em
                                [
                                    Zhang et al.~\cite{zhang2021motif}
                                    , leaf, text width=6em
                                ]
                            ]
                        ]
                    ]
                ]
                [
                    Molecule-Molecule\\ Interaction Prediction
                    [
                        \textsc{MolPair}
                        , para, text width=4em
                        [
                        Abdelaziz et al.~\cite{DBLP:journals/ws/AbdelazizFHZS17}{,}
                        CASTER~\cite{DBLP:conf/aaai/HuangXHGS20}{,}\\
                        GMPNN~\cite{nyamabo2021drug}{,}
                        MIRACLE~\cite{DBLP:conf/www/WangMCW21}
                        , leaf, text width=12em
                            [
                            \textsc{NodePair}
                            , para, text width=4em
                                [
                                KGNN~\cite{DBLP:conf/ijcai/LinQWMZ20}{,}
                                SumGNN~\cite{DBLP:journals/bioinformatics/YuHZGSX21}{,}
                                MDNN~\cite{DBLP:conf/ijcai/LyuGTLZZ21}
                                , leaf, text width=14em
                                ]
                            ]
                        ]
                    ]
                ]
                [
                    Molecule Generation\\ \& Optimization
                    [
                        \textsc{Mol2mol}
                        , para, text width=4em
                        [
                        Gomez et al.~\cite{gomez2018automatic}{,}
                        Chemformer~\cite{DBLP:journals/mlst/IrwinDHB22}{,}\\
                        LIMO~\cite{DBLP:conf/icml/EckmannSZFGY22}{,}
                        MolGen~\cite{DBLP:journals/corr/abs-2301-11259}
                        , leaf, text width=12em
                            [
                            \textsc{ActPred}
                            , para, text width=4em
                                [
                                Li et al.~\cite{DBLP:journals/corr/abs-1803-03324}{,}
                                JT-VAE\cite{DBLP:conf/icml/JinBJ18}{,}
                                GCPN~\cite{DBLP:conf/nips/YouLYPL18}{,}
                                Bradshaw et al.~\cite{DBLP:conf/nips/BradshawPKSH19}{,}\\
                                Korovina et al.~\cite{DBLP:conf/aistats/KorovinaXKNPSX20}{,}
                                Gottipati et al.~\cite{DBLP:conf/icml/GottipatiSNPWLL20}
                                , leaf, text width=20em
                                ]
                            ]
                        ]
                    ]
                ]
                [
                    Reaction \\ \& Retrosynthesis
                    [
                        \textsc{MolPair} / \textsc{Pred}
                        , para, text width=6em
                        [
                        Coley et al.~\cite{coley2017computer}{,}
                        Segler et al.~\cite{segler2017neural}{,}\\
                        Coley et al.~\cite{coley2017prediction}{,}
                        Jin et al.~\cite{DBLP:conf/nips/JinCBJ17}
                        , leaf, text width=11em
                            [
                            \textsc{NodePair} \& \textsc{Pred}
                            , para, text width=7em
                                [
                                Jin et al.~\cite{DBLP:conf/nips/JinCBJ17}
                                , leaf, text width=5em
                                ]
                            ]
                        ]
                    ]
                    [
                        \textsc{Mol2Mol}
                        , para, text width=4em
                        [
                        Liu et al.~\cite{liu2017retrosynthetic}{,}
                        SCROP~\cite{DBLP:journals/jcisd/ZhengRZXY20}
                        , leaf, text width=9em
                            [
                            \textsc{NodePair} \& \textsc{Mol2Mol} /\textsc{ActPred}
                            , para, text width=13em
                                [
                                RetroXpert~\cite{DBLP:conf/nips/YanDZZY0H20}{,}
                                G2G~\cite{DBLP:conf/icml/ShiXG0T20}
                                , leaf, text width=9em
                                ]
                            ]
                        ]
                    ]
                ]
                [ All Molecular Tasks
                    [
                        \textsc{Llm},
                        , para, text width=2em
                        [
                        Mol-Instructions~\cite{DBLP:journals/corr/abs-2306-08018}{,} ChemCrow~\cite{bran2023chemcrow}{,} 
                        ChatMol~\cite{DBLP:journals/corr/abs-2306-11976}
                        , leaf, text width=16em
                        ]
                    ]
                ]
            ]
        \end{forest}
    }
    \caption{Taxonomy of knowledge-informed paradigm transfer on molecular learning.}
    \label{fig:paradigm transfer}
   \vspace{-0.5cm}
\end{figure*}
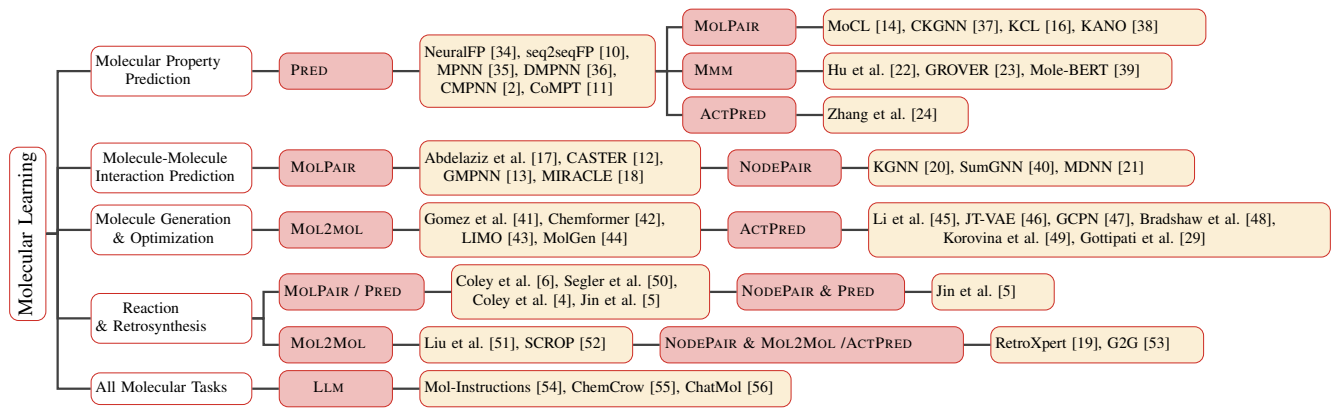

\subsection{Molecular Property Prediction}
Accurate prediction of molecular properties forms the backbone of many fundamental tasks in the chemical and pharmaceutical industries.

\subsubsection{\textsc{Pred}}
The advent of deep learning has enabled more effective solutions to the conventional molecular property prediction task under the \textsc{Pred} paradigm using a supervised approach. 
The input molecule $\mathcal{X}$ is first encoded, and then fed into a classifier to predict its property.
Previous works used RNN-based models to generate molecular representations from SMILES strings~\cite{DBLP:conf/nips/DuvenaudMABHAA15,DBLP:conf/bcb/0005WZH17}. 
To incorporate topology information in molecular graphs, MPNN~\cite{DBLP:conf/icml/GilmerSRVD17} and its variants DMPNN~\cite{DBLP:journals/jcisd/YangSJCEGGHKMPS19}, CMPNN~\cite{DBLP:conf/ijcai/SongZNFLY20}, CoMPT~\cite{DBLP:journals/corr/abs-2107-08773} utilize node and edge attributes during message passing. However, these methods still face hurdles to overcome. The scarcity of sufficient labeled data, due to the time-consuming and costly nature of lab-based data annotation, presents a major challenge. Additionally, the vast diversity of chemical molecules often leads to difficulties in generalizing models to unseen cases, which further hampers their practical application.

To mitigate these issues, recent works have embraced the self-supervised learning (SSL) approach~\cite{DBLP:conf/nips/LiuDL19,DBLP:conf/iclr/HuLGZLPL20}. 
To introduce additional chemical constraints and enhance model interpretability, domain knowledge is incorporated in pre-training a model on a large corpus of unlabeled molecular data.
Subsequently, fine-tuning is performed on downstream tasks using a limited set of labeled samples.
With the assistance of knowledge-informed pre-train models, various paradigms have demonstrated promising results in molecular property prediction. 
In the context of SSL methods, the pretext tasks can be broadly categorized into two distinct patterns: contrastive methods and predictive methods.

\subsubsection{From \textsc{Pred} to \textsc{MolPair}}
Contrastive methods construct pairwise augmented data to perform \textsc{MolPair}-based paradigm, which aims to make positive pairs closer and negative pairs further apart. An important aspect is generating diverse and informative views from each data instance. 
A bioisosteres substitution method is proposed in~\cite{DBLP:journals/corr/abs-2106-04509}, utilizing domain knowledge to guide the graph augmentation process, introducing variation while preserving molecular semantics. 
KCL~\cite{DBLP:conf/aaai/FangZYZD0Q0FC22} builds a Chemical Element KG based on the Periodic Table of Elements and augments the original molecular graph using this knowledge. This helps to establish connections between atoms that share common characteristics but are not directly linked by bonds. 
Expanding on it, KANO~\cite{fang2023knowledge} further introduces functional group prompts to stimulate the pre-trained model.
In addition to leveraging domain knowledge to enhance the graph augmentation process, CKGNN~\cite{DBLP:journals/corr/abs-2103-13047} selects positive pairs based on fingerprint similarity, aiming to learn from chemical space and incorporate domain knowledge into the output embedding space.

On the other hand, predictive methods use intrinsic properties of the data to construct prediction tasks, which can further be divided into the following two categories based on their construction techniques. 

\subsubsection{From \textsc{Pred} to \textsc{Mmm}}
Hu et al.~\cite{DBLP:conf/iclr/HuLGZLPL20} first implement the \textsc{Mmm} paradigm for molecular property prediction, and Mole-BERT~\cite{DBLP:conf/iclr/XiaZHG0LLL23} introduces a masked atoms modeling pretext task.
This involves predicting the contextual graph structures and masked node/edge attributes of a masked molecule, with the goal of capturing inherent domain knowledge through the analysis of attribute distributions. 
Instead of predicting the atom/bond type, GROVER~\cite{DBLP:conf/nips/RongBXX0HH20} randomly masks a local subgraph of the target node/edge and predicts the contextual property based on node embeddings.

\subsubsection{From \textsc{Pred} to \textsc{ActPred}}
Zhang et al.~\cite{zhang2021motif} propose a novel motif generation task that leverages the \textsc{ActPred} paradigm, aiming to exploit the data distribution of graph motifs. 
Motifs can be defined as significant subgraph patterns that frequently occur and contain semantic meaning that is indicative of the whole graph's characteristics. Motifs in molecules, such as functional groups, have a significant impact on molecular properties, emphasizing the importance of incorporating this domain knowledge. 
Following the \textsc{ActPred} paradigm, the GNN is trained to predict the next action at each time step $t$, which could be to generate a child motif with a specific label for the current motif. 

By leveraging domain knowledge in pre-training models, molecular prediction tasks can benefit from incorporating various paradigms. Domain knowledge seamlessly integrates the \textsc{MolPair}, \textsc{Mmm}, and \textsc{ActPred} paradigms into the pre-train model, narrowing the gap between them and enhancing the performance of the model in the presence of limited training data.

\subsection{Molecule-Molecule Interaction Prediction}
Molecule-molecule interaction prediction is a challenging problem in pharmacology and clinical application, as effectively identifying potential molecular interactions during clinical trials is critical for patients and society. 

\subsubsection{\textsc{MolPair}}
Most traditional molecule-molecule interaction prediction methods are usually modeled in the \textsc{MolPair} paradigm, where the two input molecules $(\mathcal{X}_a, \mathcal{X}_b)$ are encoded and interact, followed by a classifier to predict the relationship between them. Abdelaziz et al.~\cite{DBLP:journals/ws/AbdelazizFHZS17} integrate multiple drug features to calculate the similarity among drugs, and then accurately predict drug-drug interaction (DDI) based on the fused similarity. 
CASTER~\cite{DBLP:conf/aaai/HuangXHGS20} develops an end-to-end dictionary learning framework for predicting DDI with chemical structures of drugs. 
GMPNN~\cite{nyamabo2021drug} introduces a message passing network that learns chemical substructures with different sizes and shapes from the molecular graph representations of drugs for DDI prediction between a pair of drugs. 
MIRACLE~\cite{DBLP:conf/www/WangMCW21} learns drug embeddings from a multi-view graph perspective by designing a graph-based contrastive learning framework.
Although these methods have achieved strong performance, they model molecule-molecule interaction as an independent data sample and do not take into account their related correlations.

\subsubsection{From \textsc{MolPair} to \textsc{NodePair}}
Owing to the ubiquity of KG, an influx of research on it has been triggered. 
KG is a powerful tool for providing structured information on multiple entities and semantic relations associated with entities. 
Therefore, in this task, with the domain knowledge provided by these published biomedical KGs, the \textsc{NodePair} paradigm is naturally adopted. 
The main idea is to treat molecules as nodes in the KG and predict whether there is an association between the target node pairs based on the relations between these entities. 
KGNN~\cite{DBLP:conf/ijcai/LinQWMZ20} constructs a KG from raw data in the DrugBank dataset, interlinking data containing multiple types of biological drug-related entities (e.g., drug, protein, transporters, and disease).
Then, a GNN is used to explore the topology of drugs in the KG for potential DDI prediction. 
SumGNN~\cite{DBLP:journals/bioinformatics/YuHZGSX21} leverages a knowledge summarization GNN to efficiently anchor on a subgraph of potential biomedical entities that are close to the pairs in the KG, and then integrate diverse sources of external biomedical knowledge to generate a sufficient drug pair representation. 
In addition to the knowledge from KG, MDNN~\cite{DBLP:conf/ijcai/LyuGTLZZ21} also collects heterogeneous features (HF) to calculate the drug similarity between DDI events. 
A two-pathway framework, including a KG-based pathway and an HF-based pathway, is designed to obtain drug multimodal representations, where the former uses GNN to encode the KG and the latter extracts predictive information from different modalities.

\subsection{Molecule Generation and Optimization}
The main challenge of molecule generation and optimization is to find target molecules with desired chemical properties.

\subsubsection{\textsc{Mol2Mol}}
The past decade has witnessed significant advances in the domain of machine learning for molecule generation, especially in deep generative models.
The \textsc{Mol2Mol} paradigm has emerged as a promising strategy for designing novel molecules with desirable properties, which first learns to represent molecules $\mathcal{X}$ in a continuous manner to facilitate the prediction and optimization of their properties, and then map an optimized continuous representation back into a molecule $\mathcal{X}'$ with improved properties. 
Prior works in this line formulate this task as a string generation problem in an attempt to bypass generating graphs. 
Specifically, these models~\cite{gomez2018automatic,DBLP:journals/mlst/IrwinDHB22} start by generating SMILES, a linear string notation used in chemistry to describe molecular structures, which can be translated into molecules via deterministic mappings (e.g., using RDKit). 
The fragility of SMILES representations presents a significant hurdle in optimizing molecular properties, as there is no mechanism in place to ensure the validity of generated strings based on syntax and physical principles. 
However, this drawback has been effectively addressed with the advent of SELFIES, a 100\% robust molecular language that guarantees every conceivable combination of symbols in the alphabet corresponds to a chemically sound molecular structure.
Recent works, such as LIMO~\cite{DBLP:conf/icml/EckmannSZFGY22} and MolGen~\cite{DBLP:journals/corr/abs-2301-11259}, have begun to utilize variational autoencoders (VAEs) and Bidirectional and Auto-Regressive Transformers (BART) to model SELFIES for molecule generation.

\subsubsection{From \textsc{Mol2Mol} to \textsc{ActPred}}
Since essential chemical properties such as molecule validity are easier to express on graphs, manipulating directly on graphs can improve the generative modeling of valid chemical structures. Through such operations, molecule generation and optimization tasks can also be solved with the help of \textsc{ActPred}. 

One branch of graph-based methodologies relies on utilizing deep generative models. 
Li et al.~\cite{DBLP:journals/corr/abs-1803-03324} employ the \textsc{ActPred} paradigm by producing molecular graphs in a manner that resembles graph grammars. 
During derivation, the model is asked to predict the next action, i.e., whether a new structure (e.g., a new atom or a new bond) should be added to the existing graph, while the probability of this addition event depends on the history of the graph derivation. 
However, atom-by-atom construction of molecules would compel the model to generate chemically unsound intermediates, and thus, deferring validation until a complete graph is formed, such deep generative model for graphs is not ideal for molecules. 
Instead, JT-VAE\cite{DBLP:conf/icml/JinBJ18} resolves this issue by employing valid subgraphs as components, in another form of \textsc{ActPred}.
t first converts a molecular graph into a junction tree by contracting specific vertices into a single node and then represents the tree in a continuous manner. 
This phase is guided by domain knowledge, where subgraph components refer to valid chemical substructures that are automatically extracted using tree decomposition. The tree decoder traverses the entire tree and generates subgraph components in depth-first order. 
At every visited motif, the decoder is prompted to forecast the next action, such as whether to construct a children subgraph component for the current component.

Another line of graph-based molecular generation methods involves reinforcement learning (RL), which trains an agent to optimize the properties of generated molecules, naturally fitting the \textsc{ActPred} paradigm. 
The RL-based methods are mainly divided into two categories based on different optimization schemes.
The first category is RL-based graph modification, which formulates the graph generation problem as learning an RL agent that iteratively adds substructures and edges to the molecular graph in a chemistry-aware environment.
GCPN~\cite{DBLP:conf/nips/YouLYPL18} predicts the action of the bond addition, and is trained via policy gradient to optimize a reward composed of molecular property objectives and adversarial loss provided by a GCN-based discriminator trained jointly on example molecules. 
Domain knowledge is applied to constrain the model to ensure that the generated graph at each time step satisfies a set of hard constraints described by chemical valency.
While these approaches have yielded promising results, they do not guarantee synthetic feasibility.  To accomplish this, Policy Gradient for Forward Synthesis (PGFS) is proposed as a second category of RL-based methods. 
PGFS treats the generation of a molecular structure as a sequential decision-making process for the selection of reactant molecules and reaction transformations in a linear synthetic sequence. 
At each time step $t$, a reactant $\mathcal{R}_t$ is selected to react with the existing molecule $\mathcal{X}_{t-1}$ to yield the product $\mathcal{X}_{t}$, which is the molecule for the next time step. $\mathcal{X}_{t-1}$ is considered as the current state $c_{t-1}$ and the agent chooses an action $a_t$ which is further used to compute $\mathcal{R}_t$. The product $\mathcal{X}_t$ is determined by the environment based on the two reactants ($\mathcal{R}_t$ and $\mathcal{X}_{t-1}$).
Since the action space is very large with over a hundred thousand possible second reactants, incorporating domain knowledge is necessary to enumerate hypothetical product molecules accessible from libraries of available starting materials. 
Recent works, such as ~\cite{DBLP:conf/nips/BradshawPKSH19} and ~\cite{DBLP:conf/aistats/KorovinaXKNPSX20}, propose using reaction prediction algorithms to constrain searches to synthetically-accessible structures. 
More recently, Gottipati et al.~\cite{DBLP:conf/icml/GottipatiSNPWLL20} introduce an intermediate action to reduce the space of reactants by choosing a reaction template, which specifies a molecular subgraph pattern to which it can be applied and the corresponding graph transformation.

\subsection{Reaction and Retrosynthesis Prediction}
One of the fundamental problems in organic synthesis is the prediction of which products form as a result of a chemical reaction. Another is to identify a series of chemical transformations for synthesizing a target molecule. The main computational hurdle in this field is exploring the vast combinatorial space of reactions that can generate target molecules. Two predominant methods have emerged to tackle this issue - template-based and template-free approaches.

\subsubsection{\textsc{MolPair} / \textsc{Pred}}
Template-based methods match a target molecule against a large set of templates, which are molecular subgraph patterns that highlight changes during a chemical reaction. Coley et al.~\cite{coley2017computer} rank templates based on molecular similarity to precedent reactions, conforming to the routine of \textsc{MolPair}. 
Segler et al.~\cite{segler2017neural} apply the \textsc{Pred} paradigm and learn a deep neural network over the template set to predict the probability of all template rules. 
Since multiple templates can match a set of reactions, several neural models are trained to filter candidate products (reactants) using the \textsc{Pred} paradigm ~\cite{coley2017prediction,DBLP:conf/nips/JinCBJ17}. 
Generally, template-based methods usually combine the two paradigms mentioned above, but they suffer from coverage and scalability issues.
A large number of templates are required to guarantee that at least one can reconstruct the correct product (reactants).  
Despite their interpretability, template-based techniques are expensive and are not effective in generalizing to new reactions because templates are either manually created by experts or derived from reaction databases.

\subsubsection{\textsc{Mol2Mol}}
Template-free methods circumvent the use of templates by directly mapping SMILES strings of products (reactants) to reactants (products) in the form of \textsc{Mol2Mol}.
Liu et al.~\cite{liu2017retrosynthetic} first treat retrosynthesis as a machine translation process, converting SMILES by LSTM.
SCROP~\cite{DBLP:journals/jcisd/ZhengRZXY20} adds a grammar corrector to the traditional Transformer to resolve the grammatically invalid output. 
However, SMILES strings easily ignore the rich structural information in molecular graphs and cannot reasonably explain the validity of the generated SMILES.

To improve the interpretability behind the predictions, another line of methods inspired by the expert experience of chemists is proposed. 
These methods perform reaction and retrosynthesis predictions in two stages. 
In the first stage, potential reaction centers are identified to obtain intermediate molecules called synthons. In the second stage, synthons are completed into reactants (products) by sequentially generating atoms or SMILES strings. The reaction center is defined as a set of bonds that are broken during the reaction (retrosynthesis), and synthons can be obtained by splitting a target molecule according to the reaction center.

\subsubsection{to \textsc{NodePair} \& \textsc{Pred}}
For the reaction prediction task, a combination of the \textsc{NodePair} and \textsc{Pred} paradigms can be leveraged to model the aforementioned two-step strategy. 
Jin et al.~\cite{DBLP:conf/nips/JinCBJ17} train a neural network operating on the reactant graph to predict a reactivity score for each pair of atoms with the \textsc{NodePair} paradigm. 
Reaction centers are then selected by picking a small number of atom pairs with the highest reactivity scores. 
Subsequently, viable candidate products are generated by enumerating possible bond configurations between atoms within the reaction center, subject to chemical constraints. 
Finally, the \textsc{Pred} paradigm is adopted to train another neural network that ranks these candidates in such a way that the correct reaction outcome is ranked highest.

\subsubsection{to \textsc{NodePair} \& \textsc{Mol2Mol} /\textsc{ActPred}}
As for the counterpart of retrosynthesis, \textsc{NodePair} in conjunction with \textsc{Mol2Mol} or \textsc{ActPred} can be applied to tackle this problem. 
RetroXpert~\cite{DBLP:conf/nips/YanDZZY0H20} proposes to identify the potential reaction centers within the target molecule using a novel GNN, and then employs a Transformer-based mechanism to translate SMILES representations of synthon graphs into reactants by the \textsc{Mol2Mol} paradigm. 
On the other hand, G2G~\cite{DBLP:conf/icml/ShiXG0T20} completes the second step via a series of graph transformations, where a synthon can be translated into a reactant by sampling action sequences from the distribution. Consistent with \textsc{ActPred}, they first autoregressively generate a sequence of graph transformation actions (e.g., edge labeling, node selection), and then apply them on the initial synthon graph. 

\subsection{Unified Resolution for All Tasks}
While smaller models have excelled in specific tasks, the advent of LLMs has subtly revolutionized the role of human language. 
The essence lies in unleashing the potential of LLMs to benefit humanity. 
This empowers researchers to directly articulate their study requirements, yielding desired outcomes without delving into the intricate processes of molecular generation or prediction. 
Such an undertaking emerges as a paramount imperative in addressing the practical exigencies of the real world.

Some pioneering efforts initially aimed to generate specific molecular structures from text-based descriptive text~\cite{DBLP:conf/emnlp/EdwardsLRHCJ22,DBLP:journals/corr/abs-2212-10789,DBLP:journals/corr/abs-2209-05481}. Recent work has expanded the application of text-molecule interactions to a broader range of tasks.
Mol-Instructions~\cite{DBLP:journals/corr/abs-2306-08018} has curated an instructional dataset encompassing all aforementioned molecular tasks, streamlining the structural knowledge of molecules and harmonizing molecular language with human language.
ChemCrow~\cite{bran2023chemcrow} integrates 13 expert-designed chemical tools, enabling large language models to invoke these tools by following human instructions. Ultimately, LLMs determine tool selection and output final results through an automated iterative thinking process.
ChatMol~\cite{DBLP:journals/corr/abs-2306-11976} employs a novel interactive paradigm using natural language to describe and edit target molecules. In multi-turn conversations, the model is tasked with generating readable property descriptions or modified molecules that fulfill specified requirements.

\section{Conclusion and Discussion}\label{conclusion}
This paper offers a comprehensive review of molecular learning, with a focus on the knowledge-informed paradigm transfer approach. It provides an overview of the various paradigms and their technical solutions, as well as a summary of the external domain knowledge used to guide the transfer process for each molecular learning task.

\subsection{Trends Analysis}
The challenge of molecular learning lies in handling vast amounts of observational data that are inherently difficult to interpret.  Unveiling interpretable information and knowledge from such data remains an arduous task for machine learning models, which often suffer from poor generalization performance and a limited ability to generate chemically valid predictions. 
To overcome these limitations, knowledge-informed molecular learning has emerged as a promising approach to integrate fundamental domain knowledge and enhance learning performance.
In recent years, paradigm transfer has gained significant traction, with various forms of domain knowledge guiding the learning process. Accumulated facts and rules distilled by experts, as well as the insights of chemists, are valuable sources of knowledge.
To fully harness the power of domain knowledge, various molecular learning tasks have been reformulated into more flexible and versatile paradigms.
Moreover, the transfer of ideas and domain knowledge across different tasks has proved to be highly effective. By exploiting the regularities and patterns observed in one task to improve the performance of models in another, researchers can achieve remarkable results. 
This trend is expected to continue as domain knowledge behind various tasks complements each other.

\subsection{Future Directions}
With the emergence of large-scale models, a unified framework of interacting with natural language has gradually encompassed various tasks, offering enhanced practicality and applicability. One promising direction involves a deeper integration of these expansive language models with specialized chemical and biological tools, leveraging their processing and generative capabilities to the fullest. Moreover, external feedback mechanisms, such as machine learning optimization and wet lab validation, also stand as pivotal avenues to enhance the precision and reliability of molecular generation. Through these avenues, we anticipate harnessing the potential of large language models more effectively in molecular research, propelling advancements in the realms of chemistry and biology.

\bibliographystyle{IEEEtran}
\bibliography{Reference}

\begin{thebibliography}{10}
\providecommand{\url}[1]{#1}
\csname url@samestyle\endcsname
\providecommand{\newblock}{\relax}
\providecommand{\bibinfo}[2]{#2}
\providecommand{\BIBentrySTDinterwordspacing}{\spaceskip=0pt\relax}
\providecommand{\BIBentryALTinterwordstretchfactor}{4}
\providecommand{\BIBentryALTinterwordspacing}{\spaceskip=\fontdimen2\font plus
\BIBentryALTinterwordstretchfactor\fontdimen3\font minus
  \fontdimen4\font\relax}
\providecommand{\BIBforeignlanguage}[2]{{%
\expandafter\ifx\csname l@#1\endcsname\relax
\typeout{** WARNING: IEEEtran.bst: No hyphenation pattern has been}%
\typeout{** loaded for the language `#1'. Using the pattern for}%
\typeout{** the default language instead.}%
\else
\language=\csname l@#1\endcsname
\fi
#2}}
\providecommand{\BIBdecl}{\relax}
\BIBdecl

\bibitem{DBLP:journals/ijautcomp/SunLQH22}
T.~Sun, X.~Liu, X.~Qiu, and X.~Huang, ``Paradigm shift in natural language
  processing,'' \emph{Int. J. Autom. Comput.}, vol.~19, no.~3, pp. 169--183,
  2022.

\bibitem{DBLP:conf/ijcai/SongZNFLY20}
Y.~Song, S.~Zheng, Z.~Niu, Z.~Fu, Y.~Lu, and Y.~Yang, ``Communicative
  representation learning on attributed molecular graphs,'' in \emph{Proc. of
  IJCAI}, 2020.

\bibitem{DBLP:journals/corr/abs-2012-11175}
P.~Li, J.~Wang, Y.~Qiao, H.~Chen, Y.~Yu, X.~Yao, P.~Gao, G.~Xie, and S.~Song,
  ``Learn molecular representations from large-scale unlabeled molecules for
  drug discovery,'' \emph{CoRR}, 2020.

\bibitem{coley2017prediction}
C.~W. Coley, R.~Barzilay, T.~S. Jaakkola, W.~H. Green, and K.~F. Jensen,
  ``Prediction of organic reaction outcomes using machine learning,'' \emph{ACS
  central science}, 2017.

\bibitem{DBLP:conf/nips/JinCBJ17}
W.~Jin, C.~W. Coley, R.~Barzilay, and T.~S. Jaakkola, ``Predicting organic
  reaction outcomes with weisfeiler-lehman network,'' in \emph{Proc. of
  NeurIPS}, 2017.

\bibitem{coley2017computer}
C.~W. Coley, L.~Rogers, W.~H. Green, and K.~F. Jensen, ``Computer-assisted
  retrosynthesis based on molecular similarity,'' \emph{ACS central science},
  2017.

\bibitem{DBLP:journals/jcisd/WeiningerWW89}
D.~Weininger, A.~Weininger, and J.~L. Weininger, ``{SMILES.} 2. algorithm for
  generation of unique {SMILES} notation,'' \emph{J. Chem. Inf. Comput. Sci.},
  1989.

\bibitem{DBLP:journals/mlst/KrennHNFA20}
M.~Krenn, F.~H{\"{a}}se, A.~Nigam, P.~Friederich, and A.~Aspuru{-}Guzik,
  ``Self-referencing embedded strings {(SELFIES):} {A} 100{\%} robust molecular
  string representation,'' \emph{Mach. Learn. Sci. Technol.}, vol.~1, no.~4, p.
  45024, 2020.

\bibitem{xiong2019pushing}
Z.~Xiong, D.~Wang, X.~Liu, F.~Zhong, X.~Wan, X.~Li, Z.~Li, X.~Luo, K.~Chen,
  H.~Jiang \emph{et~al.}, ``Pushing the boundaries of molecular representation
  for drug discovery with the graph attention mechanism,'' \emph{Journal of
  medicinal chemistry}, 2019.

\bibitem{DBLP:conf/bcb/0005WZH17}
Z.~Xu, S.~Wang, F.~Zhu, and J.~Huang, ``Seq2seq fingerprint: An unsupervised
  deep molecular embedding for drug discovery,'' in \emph{Proceedings of the
  8th ACM International Conference on Bioinformatics, Computational Biology,
  and Health Informatics, BCB 2017, Boston, MA, USA, August 20-23, 2017}, 2017.

\bibitem{DBLP:journals/corr/abs-2107-08773}
J.~Chen, S.~Zheng, Y.~Song, J.~Rao, and Y.~Yang, ``Learning attributed graph
  representations with communicative message passing transformer,''
  \emph{CoRR}, 2021.

\bibitem{DBLP:conf/aaai/HuangXHGS20}
K.~Huang, C.~Xiao, T.~N. Hoang, L.~Glass, and J.~Sun, ``{CASTER:} predicting
  drug interactions with chemical substructure representation,'' in \emph{Proc.
  of AAAI}, 2020.

\bibitem{nyamabo2021drug}
A.~K. Nyamabo, H.~Yu, Z.~Liu, and J.-Y. Shi, ``Drug--drug interaction
  prediction with learnable size-adaptive molecular substructures,''
  \emph{Briefings in Bioinformatics}, 2021.

\bibitem{DBLP:journals/corr/abs-2106-04509}
M.~Sun, J.~Xing, H.~Wang, B.~Chen, and J.~Zhou, ``Mocl: Contrastive learning on
  molecular graphs with multi-level domain knowledge,'' \emph{CoRR}, 2021.

\bibitem{DBLP:journals/corr/abs-2102-10056}
Y.~Wang, J.~Wang, Z.~Cao, and A.~B. Farimani, ``Molclr: Molecular contrastive
  learning of representations via graph neural networks,'' \emph{CoRR}, 2021.

\bibitem{DBLP:conf/aaai/FangZYZD0Q0FC22}
Y.~Fang, Q.~Zhang, H.~Yang, X.~Zhuang, S.~Deng, W.~Zhang, M.~Qin, Z.~Chen,
  X.~Fan, and H.~Chen, ``Molecular contrastive learning with chemical element
  knowledge graph,'' in \emph{{AAAI}}.\hskip 1em plus 0.5em minus 0.4em\relax
  {AAAI} Press, 2022, pp. 3968--3976.

\bibitem{DBLP:journals/ws/AbdelazizFHZS17}
I.~Abdelaziz, A.~Fokoue, O.~Hassanzadeh, P.~Zhang, and M.~Sadoghi,
  ``Large-scale structural and textual similarity-based mining of knowledge
  graph to predict drug-drug interactions,'' \emph{J. Web Semant.}, 2017.

\bibitem{DBLP:conf/www/WangMCW21}
Y.~Wang, Y.~Min, X.~Chen, and J.~Wu, ``Multi-view graph contrastive
  representation learning for drug-drug interaction prediction,'' in
  \emph{Proc. of WWW}, 2021.

\bibitem{DBLP:conf/nips/YanDZZY0H20}
C.~Yan, Q.~Ding, P.~Zhao, S.~Zheng, J.~Yang, Y.~Yu, and J.~Huang, ``Retroxpert:
  Decompose retrosynthesis prediction like {A} chemist,'' in \emph{Proc. of
  NeurIPS}, 2020.

\bibitem{DBLP:conf/ijcai/LinQWMZ20}
X.~Lin, Z.~Quan, Z.~Wang, T.~Ma, and X.~Zeng, ``{KGNN:} knowledge graph neural
  network for drug-drug interaction prediction,'' in \emph{Proc. of IJCAI},
  2020.

\bibitem{DBLP:conf/ijcai/LyuGTLZZ21}
T.~Lyu, J.~Gao, L.~Tian, Z.~Li, P.~Zhang, and J.~Zhang, ``{MDNN:} {A}
  multimodal deep neural network for predicting drug-drug interaction events,''
  in \emph{Proc. of IJCAI}, 2021.

\bibitem{DBLP:conf/iclr/HuLGZLPL20}
W.~Hu, B.~Liu, J.~Gomes, M.~Zitnik, P.~Liang, V.~S. Pande, and J.~Leskovec,
  ``Strategies for pre-training graph neural networks,'' in \emph{Proc. of
  ICLR}, 2020.

\bibitem{DBLP:conf/nips/RongBXX0HH20}
Y.~Rong, Y.~Bian, T.~Xu, W.~Xie, Y.~Wei, W.~Huang, and J.~Huang,
  ``Self-supervised graph transformer on large-scale molecular data,'' in
  \emph{Proc. of NeurIPS}, 2020.

\bibitem{zhang2021motif}
Z.~Zhang, Q.~Liu, H.~Wang, C.~Lu, and C.-K. Lee, ``Motif-based graph
  self-supervised learning for molecular property prediction,'' \emph{Proc. of
  NeurIPS}, 2021.

\bibitem{DBLP:conf/iclr/DaiTDSS18}
H.~Dai, Y.~Tian, B.~Dai, S.~Skiena, and L.~Song, ``Syntax-directed variational
  autoencoder for structured data,'' in \emph{Proc. of ICLR}, 2018.

\bibitem{popova2018deep}
M.~Popova, O.~Isayev, and A.~Tropsha, ``Deep reinforcement learning for de novo
  drug design,'' \emph{Science advances}, 2018.

\bibitem{DBLP:journals/natmi/WangHWWWJLZYHCC21}
J.~Wang, C.~Hsieh, M.~Wang, X.~Wang, Z.~Wu, D.~Jiang, B.~Liao, X.~Zhang,
  B.~Yang, Q.~He, D.~Cao, X.~Chen, and T.~Hou, ``Multi-constraint molecular
  generation based on conditional transformer, knowledge distillation and
  reinforcement learning,'' \emph{Nat. Mach. Intell.}, 2021.

\bibitem{yang2021hit}
S.~Yang, D.~Hwang, S.~Lee, S.~Ryu, and S.~J. Hwang, ``Hit and lead discovery
  with explorative rl and fragment-based molecule generation,'' \emph{Proc. of
  NeurIPS}, 2021.

\bibitem{DBLP:conf/icml/GottipatiSNPWLL20}
S.~K. Gottipati, B.~Sattarov, S.~Niu, Y.~Pathak, H.~Wei, S.~Liu, S.~Blackburn,
  K.~M.~J. Thomas, C.~W. Coley, J.~Tang, S.~Chandar, and Y.~Bengio, ``Learning
  to navigate the synthetically accessible chemical space using reinforcement
  learning,'' in \emph{Proc. of ICML}, 2020.

\bibitem{DBLP:journals/corr/abs-2303-08774}
OpenAI, ``{GPT-4} technical report,'' \emph{CoRR}, vol. abs/2303.08774, 2023.

\bibitem{DBLP:journals/corr/abs-2302-13971}
H.~Touvron, T.~Lavril, G.~Izacard, X.~Martinet, M.~Lachaux, T.~Lacroix,
  B.~Rozi{\`{e}}re, N.~Goyal, E.~Hambro, F.~Azhar, A.~Rodriguez, A.~Joulin,
  E.~Grave, and G.~Lample, ``Llama: Open and efficient foundation language
  models,'' \emph{CoRR}, vol. abs/2302.13971, 2023.

\bibitem{DBLP:conf/iclr/WeiBZGYLDDL22}
J.~Wei, M.~Bosma, V.~Y. Zhao, K.~Guu, A.~W. Yu, B.~Lester, N.~Du, A.~M. Dai,
  and Q.~V. Le, ``Finetuned language models are zero-shot learners,'' in
  \emph{{ICLR}}.\hskip 1em plus 0.5em minus 0.4em\relax OpenReview.net, 2022.

\bibitem{DBLP:journals/corr/abs-2210-02414}
A.~Zeng, X.~Liu, Z.~Du, Z.~Wang, H.~Lai, M.~Ding, Z.~Yang, Y.~Xu, W.~Zheng,
  X.~Xia, W.~L. Tam, Z.~Ma, Y.~Xue, J.~Zhai, W.~Chen, P.~Zhang, Y.~Dong, and
  J.~Tang, ``{GLM-130B:} an open bilingual pre-trained model,'' \emph{CoRR},
  vol. abs/2210.02414, 2022.

\bibitem{DBLP:conf/nips/DuvenaudMABHAA15}
D.~Duvenaud, D.~Maclaurin, J.~Aguilera{-}Iparraguirre,
  R.~G{\'{o}}mez{-}Bombarelli, T.~Hirzel, A.~Aspuru{-}Guzik, and R.~P. Adams,
  ``Convolutional networks on graphs for learning molecular fingerprints,'' in
  \emph{Proc. of NeurIPS}, 2015.

\bibitem{DBLP:conf/icml/GilmerSRVD17}
J.~Gilmer, S.~S. Schoenholz, P.~F. Riley, O.~Vinyals, and G.~E. Dahl, ``Neural
  message passing for quantum chemistry,'' in \emph{Proc. of ICML}, 2017.

\bibitem{DBLP:journals/jcisd/YangSJCEGGHKMPS19}
K.~Yang, K.~Swanson, W.~Jin, C.~W. Coley, P.~Eiden, H.~Gao, A.~Guzman{-}Perez,
  T.~Hopper, B.~Kelley, M.~Mathea, A.~Palmer, V.~Settels, T.~S. Jaakkola, K.~F.
  Jensen, and R.~Barzilay, ``Analyzing learned molecular representations for
  property prediction,'' \emph{J. Chem. Inf. Model.}, 2019.

\bibitem{DBLP:journals/corr/abs-2103-13047}
Y.~Fang, H.~Yang, X.~Zhuang, X.~Shao, X.~Fan, and H.~Chen, ``Knowledge-aware
  contrastive molecular graph learning,'' \emph{CoRR}, 2021.

\bibitem{fang2023knowledge}
Y.~Fang, Q.~Zhang, N.~Zhang, Z.~Chen, X.~Zhuang, X.~Shao, X.~Fan, and H.~Chen,
  ``Knowledge graph-enhanced molecular contrastive learning with functional
  prompt,'' \emph{Nature Machine Intelligence}, pp. 1--12, 2023.

\bibitem{DBLP:conf/iclr/XiaZHG0LLL23}
J.~Xia, C.~Zhao, B.~Hu, Z.~Gao, C.~Tan, Y.~Liu, S.~Li, and S.~Z. Li,
  ``Mole-bert: Rethinking pre-training graph neural networks for molecules,''
  in \emph{{ICLR}}.\hskip 1em plus 0.5em minus 0.4em\relax OpenReview.net,
  2023.

\bibitem{DBLP:journals/bioinformatics/YuHZGSX21}
Y.~Yu, K.~Huang, C.~Zhang, L.~M. Glass, J.~Sun, and C.~Xiao, ``Sumgnn:
  multi-typed drug interaction prediction via efficient knowledge graph
  summarization,'' \emph{Bioinform.}, 2021.

\bibitem{gomez2018automatic}
R.~G{\'o}mez-Bombarelli, J.~N. Wei, D.~Duvenaud, J.~M. Hern{\'a}ndez-Lobato,
  B.~S{\'a}nchez-Lengeling, D.~Sheberla, J.~Aguilera-Iparraguirre, T.~D.
  Hirzel, R.~P. Adams, and A.~Aspuru-Guzik, ``Automatic chemical design using a
  data-driven continuous representation of molecules,'' \emph{ACS central
  science}, 2018.

\bibitem{DBLP:journals/mlst/IrwinDHB22}
R.~Irwin, S.~Dimitriadis, J.~He, and E.~J. Bjerrum, ``Chemformer: a pre-trained
  transformer for computational chemistry,'' \emph{Mach. Learn. Sci. Technol.},
  vol.~3, no.~1, p. 15022, 2022.

\bibitem{DBLP:conf/icml/EckmannSZFGY22}
P.~Eckmann, K.~Sun, B.~Zhao, M.~Feng, M.~K. Gilson, and R.~Yu, ``{LIMO:} latent
  inceptionism for targeted molecule generation,'' in \emph{{ICML}}, ser.
  Proceedings of Machine Learning Research, vol. 162.\hskip 1em plus 0.5em
  minus 0.4em\relax {PMLR}, 2022, pp. 5777--5792.

\bibitem{DBLP:journals/corr/abs-2301-11259}
Y.~Fang, N.~Zhang, Z.~Chen, X.~Fan, and H.~Chen, ``Molecular language model as
  multi-task generator,'' \emph{CoRR}, vol. abs/2301.11259, 2023.

\bibitem{DBLP:journals/corr/abs-1803-03324}
Y.~Li, O.~Vinyals, C.~Dyer, R.~Pascanu, and P.~W. Battaglia, ``Learning deep
  generative models of graphs,'' \emph{CoRR}, 2018.

\bibitem{DBLP:conf/icml/JinBJ18}
W.~Jin, R.~Barzilay, and T.~S. Jaakkola, ``Junction tree variational
  autoencoder for molecular graph generation,'' in \emph{Proc. of ICML}, 2018.

\bibitem{DBLP:conf/nips/YouLYPL18}
J.~You, B.~Liu, Z.~Ying, V.~S. Pande, and J.~Leskovec, ``Graph convolutional
  policy network for goal-directed molecular graph generation,'' in \emph{Proc.
  of NeurIPS}, 2018.

\bibitem{DBLP:conf/nips/BradshawPKSH19}
J.~Bradshaw, B.~Paige, M.~J. Kusner, M.~H.~S. Segler, and J.~M.
  Hern{\'{a}}ndez{-}Lobato, ``A model to search for synthesizable molecules,''
  in \emph{Proc. of NeurIPS}, 2019.

\bibitem{DBLP:conf/aistats/KorovinaXKNPSX20}
K.~Korovina, S.~Xu, K.~Kandasamy, W.~Neiswanger, B.~P{\'{o}}czos, J.~Schneider,
  and E.~P. Xing, ``Chembo: Bayesian optimization of small organic molecules
  with synthesizable recommendations,'' in \emph{Proc. of AISTATS}, 2020.

\bibitem{segler2017neural}
M.~H. Segler and M.~P. Waller, ``Neural-symbolic machine learning for
  retrosynthesis and reaction prediction,'' \emph{Chemistry--A European
  Journal}, 2017.

\bibitem{liu2017retrosynthetic}
B.~Liu, B.~Ramsundar, P.~Kawthekar, J.~Shi, J.~Gomes, Q.~Luu~Nguyen, S.~Ho,
  J.~Sloane, P.~Wender, and V.~Pande, ``Retrosynthetic reaction prediction
  using neural sequence-to-sequence models,'' \emph{ACS central science}, 2017.

\bibitem{DBLP:journals/jcisd/ZhengRZXY20}
S.~Zheng, J.~Rao, Z.~Zhang, J.~Xu, and Y.~Yang, ``Predicting retrosynthetic
  reactions using self-corrected transformer neural networks,'' \emph{J. Chem.
  Inf. Model.}, 2020.

\bibitem{DBLP:conf/icml/ShiXG0T20}
C.~Shi, M.~Xu, H.~Guo, M.~Zhang, and J.~Tang, ``A graph to graphs framework for
  retrosynthesis prediction,'' in \emph{Proc. of ICML}, 2020.

\bibitem{DBLP:journals/corr/abs-2306-08018}
Y.~Fang, X.~Liang, N.~Zhang, K.~Liu, R.~Huang, Z.~Chen, X.~Fan, and H.~Chen,
  ``Mol-instructions: {A} large-scale biomolecular instruction dataset for
  large language models,'' \emph{CoRR}, vol. abs/2306.08018, 2023.

\bibitem{bran2023chemcrow}
A.~M. Bran, S.~Cox, A.~D. White, and P.~Schwaller, ``Chemcrow: Augmenting
  large-language models with chemistry tools,'' \emph{arXiv preprint
  arXiv:2304.05376}, 2023.

\bibitem{DBLP:journals/corr/abs-2306-11976}
Z.~Zeng, B.~Yin, S.~Wang, J.~Liu, C.~Yang, H.~Yao, X.~Sun, M.~Sun, G.~Xie, and
  Z.~Liu, ``Interactive molecular discovery with natural language,''
  \emph{CoRR}, vol. abs/2306.11976, 2023.

\bibitem{DBLP:conf/nips/LiuDL19}
S.~Liu, M.~F. Demirel, and Y.~Liang, ``N-gram graph: Simple unsupervised
  representation for graphs, with applications to molecules,'' in \emph{Proc.
  of NeurIPS}, 2019.

\bibitem{DBLP:conf/emnlp/EdwardsLRHCJ22}
C.~Edwards, T.~M. Lai, K.~Ros, G.~Honke, K.~Cho, and H.~Ji, ``Translation
  between molecules and natural language,'' in \emph{{EMNLP}}.\hskip 1em plus
  0.5em minus 0.4em\relax Association for Computational Linguistics, 2022, pp.
  375--413.

\bibitem{DBLP:journals/corr/abs-2212-10789}
S.~Liu, W.~Nie, C.~Wang, J.~Lu, Z.~Qiao, L.~Liu, J.~Tang, C.~Xiao, and
  A.~Anandkumar, ``Multi-modal molecule structure-text model for text-based
  retrieval and editing,'' \emph{CoRR}, vol. abs/2212.10789, 2022.

\bibitem{DBLP:journals/corr/abs-2209-05481}
B.~Su, D.~Du, Z.~Yang, Y.~Zhou, J.~Li, A.~Rao, H.~Sun, Z.~Lu, and J.~Wen, ``A
  molecular multimodal foundation model associating molecule graphs with
  natural language,'' \emph{CoRR}, vol. abs/2209.05481, 2022.

\end{thebibliography}

\end{document}